# Luganda Text-to-Speech Machine

**Irene Nandutu. Uganda Technology and Management University.**

irenenanduttu@gmail.com

**Ernest Mwebaze. Makerere University.** emwebaze@cit.ac.ug

## Abstract

In Uganda, Luganda is the most spoken native language. It is used for communication in informal as well as formal business transactions. The development of technology startups globally related to TTS has mainly been with languages like English, French, etc. These are added in TTS engines by Google, Microsoft among others, allowing developers in these regions to innovate TTS products. Luganda is not supported because the language is not built and trained on these engines. In this study, we analyzed the Luganda language structure and constructions and then proposed and developed a Luganda TTS. The system was built and trained using locally sourced Luganda language text and audio. The engine is now able to capture text and reads it aloud. We tested the accuracy using MRT and MOS. MRT and MOS tests results are quite good with MRT having better results. The results general score was 71%. This study will enhance previous solutions to NLP gaps in Uganda, as well as provide raw data such that other research in this area can take place.

## KEYWORDS

TTS-Text-to-Speech, NLP – Natural Language Processing, MRT- Modified Rhythm Test , MOS-Mean Opinion Score.

## 1 Introduction

The language Luganda is part of the Niger-Congo family of languages. These sometimes are called the bantu languages. Luganda is a widely spoken native language by the Ganda people in Uganda. It is spoken by more than five million people, making the language have the highest number of speakers who are increasingly fluent. Today, we witness advancement in Luganda documented literature that is published and documented in magazines, newspapers, etc. In 2013, Moses and Eno-Abasi emphasized that most African languages are resource limited with less or no linguistic resources like textbooks [8]. Luganda is not exceptional as a few researchers have done some work on NLP e.g. [5]. This research draws from a demand for innovative technology products on Luganda language linguistics e.g. learning systems for linguistic students to practice, speaking websites and mobile apps. These products are the reason for the need of a Luganda Text-to-Speech Machine. We used MARYTTS (Modular Architecture for Research on speech sYnthesis) engine.

## 2 Related Work

A Text-to-Speech (TTS) synthesizer is a computer based system that can read text aloud automatically. A speech synthesizer can be implemented by both hardware and software [1]. Speech synthesis compliments other language technologies such as speech recognition, which aims to convert speech into text, machine translation which converts writing or speech in one language into writing or speech in another [3]. There are seven ways we can understand speech and they are; Acoustics, phonetics, phonology, morphology, syntax, semantics and pragmatics [17]. TTS machines are divided into NLP and digital processing. NLP pcaptures the dump raw text, analyses it, process it and combine information used later for the synthesis. The speech synthesis performs phonetic transcriptions, the output is usually a sequence of phones supplied with prosody makers. The speech synthesis performs speech synthesis on the signal level, based on information from the NLP module.

## 2.1 Unit selection-based synthesis

Unit are phonemes which are units of sound speech that distinguish one utterance from another in a language, these units of sound are either vowels or consonants of a particular language. To be able to create a speech units inventory, it is required to have the speech corpus and to process a segmentation. The segmentation is the process of finding boundaries of the selected speech units in the speech data. The segmentation can be manual or automatic. The automatic segmentation is mostly used especially on large speech units' inventory. There are two mainly used methods of the automatic segmentation: a Hidden Markov Models (HMM)-based method and a Dynamic Time Warping (DTW)-based method. Because of the better consistency of segmentation the HMM-based method are preferred.

| Words | butiko | | | | | |
|---|---|---|---|---|---|---|
| Syllables | bu | | ti | | ko | |
| Phonemes | b | u | t | i | k | o |
| Triphones | <sil>-b+u | b-u+t | u-t+i | t t-i+k | i-k+o | k-o +<sil> |

Fig 1: The illustration of the segmentation of the word butiko on the speech units.

[16].In unit selection, a speech database is designed such that each unit is available in various prosodic and phonetic contexts [18]. The char (text) input is captured as input to enable the utterance as output which defines the string of phonemes (vowels and consonants) required to synthesize the text, and is annotated with prosodic features (pitch, duration and power) which specify the desired speech output.

## 2.2 Other Speech Synthesis Work Done in Uganda on Ugandan Languages

The Language speech synthesis work and research done in Uganda is mainly related to natural language processing like software localization. Not much has been done to translate Ugandan local languages using a Text-to-Speech Machine. There have been software localization ventures for the Luganda [6] and Runyakitara languages. This was done on the Google interface translating it to Runyakitara [5] and [10]. The motivation behind this is that it can pave a way in bridging the Digital Divide between countries. People are motivated to embrace and participate in technologies that are in a language they are most familiar with, mainly their mother tongue [5].

## 2.3 Conclusion

There is quite a lot of literature about Text-To-Speech machine focused on both natural language processing and speech synthesis. The researchers have found literature about Software localization in the Luganda language. However, there is no literature found about Text-To-Speech on Ugandan Languages. It is on this research gap analysis that this research has been done and the results are promising.

## 3 Methodology

This research shows a Luganda Text-to-Speech system using MARYTTS. The MARYTTS TTS (Text-To-Speech) synthesis system is a flexible and modular tool for research, development and teaching in the domain of Text-To-Speech synthesis [4]. MARYTTS [8][9][7][15][14] is an open-source project, it is written in Java and includes a number of useful tools for adding support for a new language and adding new voices. The aim of these tools is to simplify the task of building new resources for TTS, their effectiveness can be seen from the fact that when MARYTTS was born it was originally developed for the German language; nowadays it makes available voices and support for the following languages: US English, British English, German, Turkish, Russian, Telugu, etc.

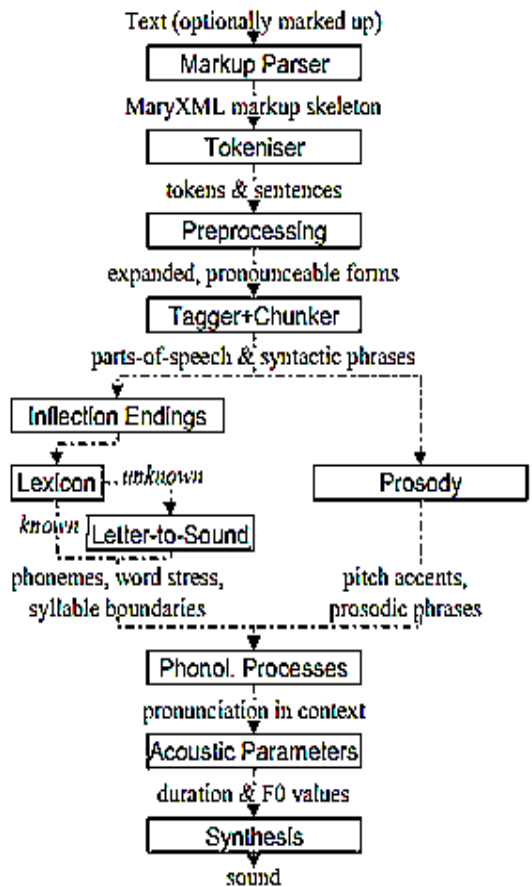

Fig 2: Architecture Structure of a MARYTTS

**3.1 Plain Text**

Plain text is the most basic, and maybe most common input format. Nothing is known about the structure or meaning of the text. The text is embedded into a MaryXML document for the following processing steps

**3.2 SABLE -annotated text and SSML-annotated text**

SABLE is a markup language for annotating texts in view of speech synthesis, and SSML is a markup language for annotating texts in view of speech synthesis. It was proposed by the W3C as a standard. Speech synthesis markup languages are useful for providing information about the structure of a document, the meaning of numbers, or the importance of words, so that this information can be appropriately expressed in speech (such as pausing in the right places, pronouncing telephone numbers appropriately, or putting emphasis on the word carrying focus). Such information may be provided by a human user or, more likely, by other processing units such as natural language generators, email processors, or HTML readers.

**3.3 Optional Markup Parser**

The MARY text-to-speech and markup-to-speech system accepts both plain text input and input marked up for speech synthesis with a speech synthesis markup language such as SABLE or SSML. . Both SABLE and SSML are transformed to MaryXML which reflects the modeling capabilities of this particular TTS system. MaryXML is based on XML

**3.4 Tokenizer**

The tokenizer cuts the text into tokens, i.e. words and punctuation marks. It uses a set of rules determined through corpus analysis to label the meaning of dots based on the surrounding contex

**3.5 Text Normalisation Module**

In the preprocessing module, organizes the input sentences into manageable lists of words. It identifies numbers, abbreviations, acronyms and idiomatic s and transforms them into full text when needed, those tokens for which spoken form does not entirely correspond to the written form are replaced by a more pronounceable form.

**3.5.1 Numbers:**

The pronunciation of numbers will highly depend on their meaning. Different number types, such as cardinal and ordinal numbers, currency amounts, or telephone numbers, must be identified as such, either from input markup or from context, and replaced by appropriate token strings.

**3.5.2. Abbreviations:**

Two main groups of abbreviations are distinguished: Those that are spelled out, such as "USA", and those that need expansion.

**3.7 Phonemisation**

The output of the phonemisation component contains the phonemic transcription. The Speech Assessment Methods Phonetic Alphabet (SAMPA) phonetic alphabet will be created for Luganda and adopted to be used for each token, as well as the source of this transcription (simple lexicon lookup, lexicon lookup with compound analysis, letter-to-sound rules, etc.).

### 3.7.1 Inflection endings:

This module deals with the ordinals and abbreviations which have been marked during preprocessing as requiring an appropriate inflection ending. The part-of-speech information added by the tagger tells whether the token is an adverb or an adjective. In addition, information about the boundaries of noun phrases has been provided by the chunker, which is relevant for adjectives.

### 3.7.2 Lexicons:

The pronunciation lexicon contains the grapheme form, a phonemic transcription, a special marking for adjectives, and the inflection information

### 3.7.3 Letter-to-sound conversion:

Unknown words that cannot be phonemised with the help of the lexicon are analyzed by a "letter-to-sound conversion" algorithm. Letter-to-Sound rules are statistically trained on the MARY lexicon.

### 3.8 Prosody Module

The prosody rules were derived through corpus analysis and are mostly based on part-of-speech and punctuation information. Some parts-of-speech, such as nouns and adjectives, always receive an accent; the other parts-of-speech are ranked hierarchically (roughly: full verbs > modal verbs > adverbs), according to their aptitude to receive an accent. This ranking comes into play where the obligatory assignment rules do not place any accent inside some intermediate phrase. According to a GToBI principle, each intermediate phrase should contain at least one pitch accent. In such a case, the token in that intermediate phrase with the highest-ranking part-of-speech receives a pitch accent. After determining the location of prosodic boundaries and pitch accents, the actual tones are assigned according to sentence type (declarative, interrogative-W, interrogative-Yes-No and exclamative). For each sentence type, pitch accent tones, intermediate phrase boundary tones and intonation phrase boundary tones are assigned. The last accent and intonation phrase tone in a sentence is usually different from the rest, in order to account for sentence-final intonation patterns.

### 3.9 Postlexical Phonological Process

Once the words are transcribed in a standard phonemic string including syllable boundaries and lexical stress on the one hand, and the prosody labels for pitch accents and prosodic phrase boundaries are assigned on the other hand, the resulting phonological representation can be restructured by a number of phonological rules. These rules operate on the basis of phonological context information such as pitch accent, word stress, the phrasal domain or, optionally, requested articulation precision.

### 3.10 Calculation of Acoustic Parameters

This module performs the translation from the symbolic to the physical domain. The output produced by this module is a list containing the individual segments with their durations as well as F0 targets. This format is compatible with the MBROLA .pho input files.

### 3.11 Synthesis

At present, MBROLA is used for synthesizing the utterance based on the output of the preceding module. Due to the modular architecture of the MARY system, any synthesis module with a similar interface could easily be employed instead or in addition.

## 4 Adding a new Language and Voice Module to MARYTTS

This chapter covers the data preparation to train the HMM models for the new voice and the new language module that was added to the MARY system.

### 4.1 Speech database

Speech database [20] is one of the components used to train unit selection models for speech synthesis, a database was recorded and for each utterance the text annotation needs to be present in the selected sentences with in the database to be able to transcribe the speech sample and to train the unit selection models properly.

### 4.1.1 Database Creation

In this stage, we select certain data from the Wikipedia dump data. Factors considered while creating the database are the purpose of the corpus, size of the corpus, the number and diversity of speakers, the environment and the style of recording, phonetic coverage. The corpus can have phonetics described in the phonetic file close to natural speech and wide enough with phonetic units (e.g. phonemes or dip-hones). Also, the selected sentences should be reasonably short and easy to read. For the purpose of this project, a small database with only one speaker was created. The data for the corpus was prepared using the MARY system's building tools. The created database corpus consists of 511 sentences.

### 4.2.2 Recording the Speech Database

Using the created text database, the corresponding speech database was recorded by one female speaker in .wav files. We recorded using redstart a MARY TTS recording tool. The format of the audio data is the following: 16 kHz sampling rate, 16-bit samples, mono wav files.

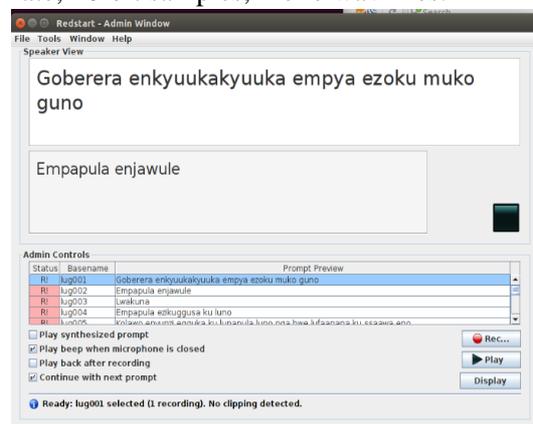

Fig 3: Restart Recording Tool

### 4.3 Integrating the New Language into MARY TTS

Adding a new language to MARY TTS follows a guideline with subsections. The whole implementation was done using Ubuntu Linux Distribution because Linux is the best platform to run the voice building software's. This research used a corei3 processor Laptop with 6GB RAM because that is want the resources enabled the researcher to have and it did a tremendous job in achieving the goal of this research, various software's were installed to help in training the unit selection models and attaining a synthetic voice.

### 4.3.1 Adding the New Language Module to MARY TTS

MARYTTS Language module uses a minimal NLP Components which helped in phonemisation (turning text into sound sequences i.e. grapheme to phoneme conversion), this is the linking point between text and speech since it converts text to allophones. Also it is were tokenization takes place and later prosodic events (accent and phrasing) from the text is performed. Three major things are put into consideration while adding a new language and these are; a complete set of phones with the articulation features, dictionary containing words with their phonetic transcriptions and list of functional words for the target language.

### 4.3.3 Adding a New Voice Module to MARY TTS

The guideline for creating a unit selection voice was followed to build a new unit selection voice. These guidelines are put into consideration when building the voice and these are; First of all data from the created database need to be prepared in the following pattern: one sentence per file, speech sound in wav file, text representation of the sentence in txt file and the names of wav and txt files have to match due to the transcription alignment calculations, e.g. files audio000.wav and audio000.txt need to contain the same sentence. Sound files have to be in directory. /wav, text files in directory. /text placed in selected working directory. These help in running the Voice Import tools together with installed programs then prepare the data for the HTS toolkit and unit selection models are trained according to the characteristics of the created database. This procedure takes a long time and the output models and trees are used for the new voice creation. In this research we build the voice using a unit selection voice using software for building the voice like praat which computes pitch markers by aligning Pitch Marks to the nearest Zero Crossings, HTS, HTK, HDecode, Speech Signal Processing Toolkit (SPTK), hts-engine among others were installed [22][23][24]. Also Edinburgh Speech Tools was installed.

### 4.5 Unit Selection Voice Creation

To build a unit selection voice creation all speech signals were in .wav format, 16 kHz sampling frequency and 16-bit sample format, all speech signals have corresponding text of what is spoken in each speech signal (.txt format). MaryTTS server must be running on local host: 59125(default).

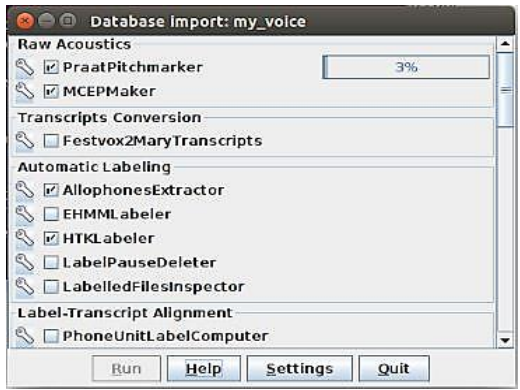
Fig 5: Voice Import Tools
Run "PraatPitchmarker" Result : voice/pm/* files
Run "MCEPMaker" Result : voice/mcep/*.mcep files
Voice compiling: Automatic Labelling
Run "AllophonesExtractor". Result: voice/prompt-allophones/*.xml
Run "HTS Labeler" Result: voice/hts/* files. This step will process several hours
Run "LabelPauseDeleter" (set threshold=10 in settings). Result: voice/lab/*.lab
Run "PhoneUnitLabelComputer". Result: voice/phonelab/*.lab
Run "TranscriptionAligner". Result: voice/allphones/*.xml
Run "FeatureSelection". Result: voice/mary/features.txt
Run "PhoneUnitFeatureComputer": Result: voice/phonefeatures/*.pfeats
Run "PhoneLabelFeatureAligner". See console output - it should display "0 problems"
Voice compiling: voice training
Run "HMMVoiceDataPreparation
Run "HMMVoiceConfigure"
Run "HMMVoiceFeatureSelection". Result: mary/hmmFeatures.txt
Run "HMMVoiceMakeData"
Run "HMMVoiceMakeVoice". This step will process several hours.
Run "HMMVoiceCompiler".
The above training was successful, the researcher published the voice to dropbox and installed it to MARYTTS by adding the link marytts-components.xml file

**MARYTTS Client**

The MARYTTS-client has a graphical user interface where we input Luganda text and process to get Luganda words, tokens, intonations, part of speech, and audio among others. The image below allows input of text and outputs audio. Program expects a text input in the top text area. Button Play sends request to the MARY server for an audio output and plays it as a speech sound.

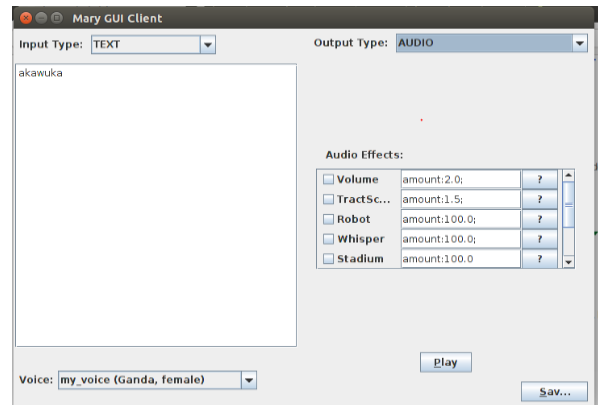
Fig 6: Luganda Text-to-Speech

### 4.7 Evaluation of the synthesized speech quality

During evaluation the intelligibility and the naturalness of the synthesized speech were tested. The intelligibility tests were evaluating how listeners understands the synthesized speech with an emphasis on perception transition sounds; the naturalness test were evaluating the speech from an overall aspect.

### 4.7.1 Modified Rhyme Test (MRT)

Modified rhyme test (MRT) was used to test the intelligibility of the synthesized speech. MRT is one of the most widely used intelligibility tests. During the test one word is played to the listener and his task is to identify the sound by looking at the word lists in their hand. The test is concentrated on consonants because the synthesis of consonants has the biggest influence on the intelligibility of the speech.

| bbiri | bibiri | ebiri | ekika | ebika | muntu |
|---|---|---|---|---|---|
| abantu | kirabo | bitabo | kintu | kaalo | akaalo |
| kolawo | obungi | misir | kabaka | engeri | nabuli |
| ebintu | kakat | emyaka | terina | kyaali | kayita |
| bamwe | magezi | kagabo | abaali | lwaffe | byaali-yimirira |
| kugoba | dungi | emyezi | kyaffe | mukasa | wabula |
| mukazi | bbanga | wo'muti | mmange | mummwe | mmanyi |
| oddire | empya | ezoku | ssatu | rujja | mumanyi |
| byaba | naffe | yaffe | ffena | abato | mwaka |
| baffe | okuba | nkumi | nange | okuva | bantu |
| ebifa | zaali | abamu | muntu | baali | twali |
| tujja | maaso | enaku | bonna | amaze | etaka |

Fig 7: Words selected for MRT test

### 4.7.2 Mean Opinion Score (MOS) Test

Mean opinion score (MOS) test was used to test the naturalness of the synthesized speech. Sentences were selected from the corpus and speech quality rated by listeners. Example of selected synthesized text; "era ndyerera ddala ennyumba ya Yerobowaamu ng'omuntu bw'ayera obusa n'okuggwaawo ne buggwaawo bwonna"

| Rating | Speech Quality | |
|---|---|---|
| 1 | Bad | No meaning understood |
| 2 | Poor | Considerable effort required |
| 3 | Fair | Moderate effort required |
| 4 | Good | Attention necessary |
| 5 | Excellent | No effort required |

Table 1: Rating for MOS test

The above scale reflects the opinion of the quality of the synthesized speech or shows what effort listeners have to spend during listening to the synthesized speech [11]

### 4.7.3 Results of Evaluation

These were evaluated inform of a questionnaire. The exercise was composed of mainly 20 students from the age of 20 years to 30 years. The overall right answers were 71%

| Rating | Speech Quality | Listening Effort | |
|---|---|---|---|
| 1 | Bad | No meaning understood | 5% |
| 2 | Poor | Considerable effort required | 20% |
| 3 | Fair | Moderate effort required | 49% |
| 4 | Good | Attention necessary | 22% |
| 5 | Excellent | No effort required | 4% |

Table 2: The MOS test evaluation results

### 4.8 Results

As a result of adding the new language module and the new voice to the MARY system, several simple programs and scripts to ease database creation and to ease system integration were created. The whole process was described step by step. The database corpus was created and the speech database was recorded. The new language module and the new voice were successfully integrated into the MARY system. A pull request is made to publish the voice in MARYTTS https://github.com/marytts/marytts/pull/792.

Meanwhile the synthesized voice can be installed from dropbox. Follow the instructions below;

- Install Java and add it to system pat
- Clone MARYTTS https://github.com/marytts/marytts or version 5.2 SNAPSHOT; this was used for testing.
- Download target folder from dropbox link which has the Luganda voice, https://goo.gl/Qi6xDo
- Unzip it and place it at the root of marytts project
- Navigate to target/marytts-5.2-SNAPSHOT/bin/marytts-server in command line. When the server is running; browse http://localhost:59125/
- https://github.com/Nandutu/luganda_data set . Use the text dataset created to test the voice. Use test/.txt folder and lg.txt to test voice. This was the little dataset used to train the voices.

Luganda corpus was developed that was used to build a Text-to-Speech machine, this machine can capture Luganda text and renders Luganda synthetic voice. Since there was no such work done in speech synthesis research, no comparison was done.